\documentclass[conference]{IEEEtran}
\usepackage{cite}
\usepackage{amsmath,amssymb,amsfonts}
\usepackage{algorithmic}
\usepackage{graphicx}
\usepackage{textcomp}
\usepackage{xcolor}

\usepackage{multirow}

\def\BibTeX{{\rm B\kern-.05em{\sc i\kern-.025em b}\kern-.08em
    T\kern-.1667em\lower.7ex\hbox{E}\kern-.125emX}}

\usepackage{subcaption}

\usepackage{tikz}
\usetikzlibrary{calc,shapes,arrows,decorations.text,positioning,patterns,matrix,chains,decorations.pathreplacing, decorations.pathmorphing}

\makeatletter
\newcommand{\linebreakand}{%
  \end{@IEEEauthorhalign}
  \hfill\mbox{}\par
  \mbox{}\hfill\begin{@IEEEauthorhalign}
}
\makeatother

\begin{document}

\title{Robust Detection of Objects under Periodic Motion with Gaussian Process Filtering
%
}

\author{\IEEEauthorblockN{Joris Guerin\IEEEauthorrefmark{1}, Anne Magaly de Paula Canuto\IEEEauthorrefmark{2} and Luiz Marcos Garcia Goncalves\IEEEauthorrefmark{1}}\\
\IEEEauthorblockA{\IEEEauthorrefmark{1}Departamento de Engenharia de Computação e Automação, \\ \IEEEauthorrefmark{2} Departamento de Informatica e Matematica Aplicada\\
Universidade Federal do Rio Grande do Norte (UFRN), Natal-RN, Brazil\\
Emails: jorisguerin.research@gmail.com, anne@dimap.ufrn.br, lmarcos@dca.ufrn.br}}

\maketitle

\thispagestyle{plain}
\pagestyle{plain}

\begin{abstract}
Object Detection (OD) is an important task in Computer Vision with many practical applications. For some use cases, OD must be done on videos, where the object of interest has a periodic motion. In this paper, we formalize the problem of periodic OD, which consists in improving the performance of an OD model in the specific case where the object of interest is repeating similar spatio-temporal trajectories with respect to the video frames. The proposed approach is based on training a Gaussian Process to model the periodic motion, and use it to filter out the erroneous predictions of the OD model. By simulating various OD models and periodic trajectories, we demonstrate that this filtering approach, which is entirely data-driven, improves the detection performance by a large margin.
\end{abstract}


\section{Introduction}
\label{sec:intro}

Object Detection (OD) is a Computer Vision task consisting in localizing members of a given object category in an image~\cite{OD_old}. In practice, this is done by predicting rectangular Bounding Boxes (BB) coordinates to encompass the studied objects in an image. Throughout the last decades, OD has been widely studied and huge progress has been made with the appearance of deep learning based approaches~\cite{fastrcnn, fasterrcnn, yolo, ssd, yolov3, detectors}. A complete survey of the OD literature is beyond of the scope of this paper. Nonetheless, two recent surveys can be found in the literature~\cite{survey_OD_deep1, survey_OD_deep2}.

\begin{figure}[!ht]
\centering
    \begin{subfigure}{.48\textwidth}
    \centering
    \includegraphics[width=0.8\textwidth]{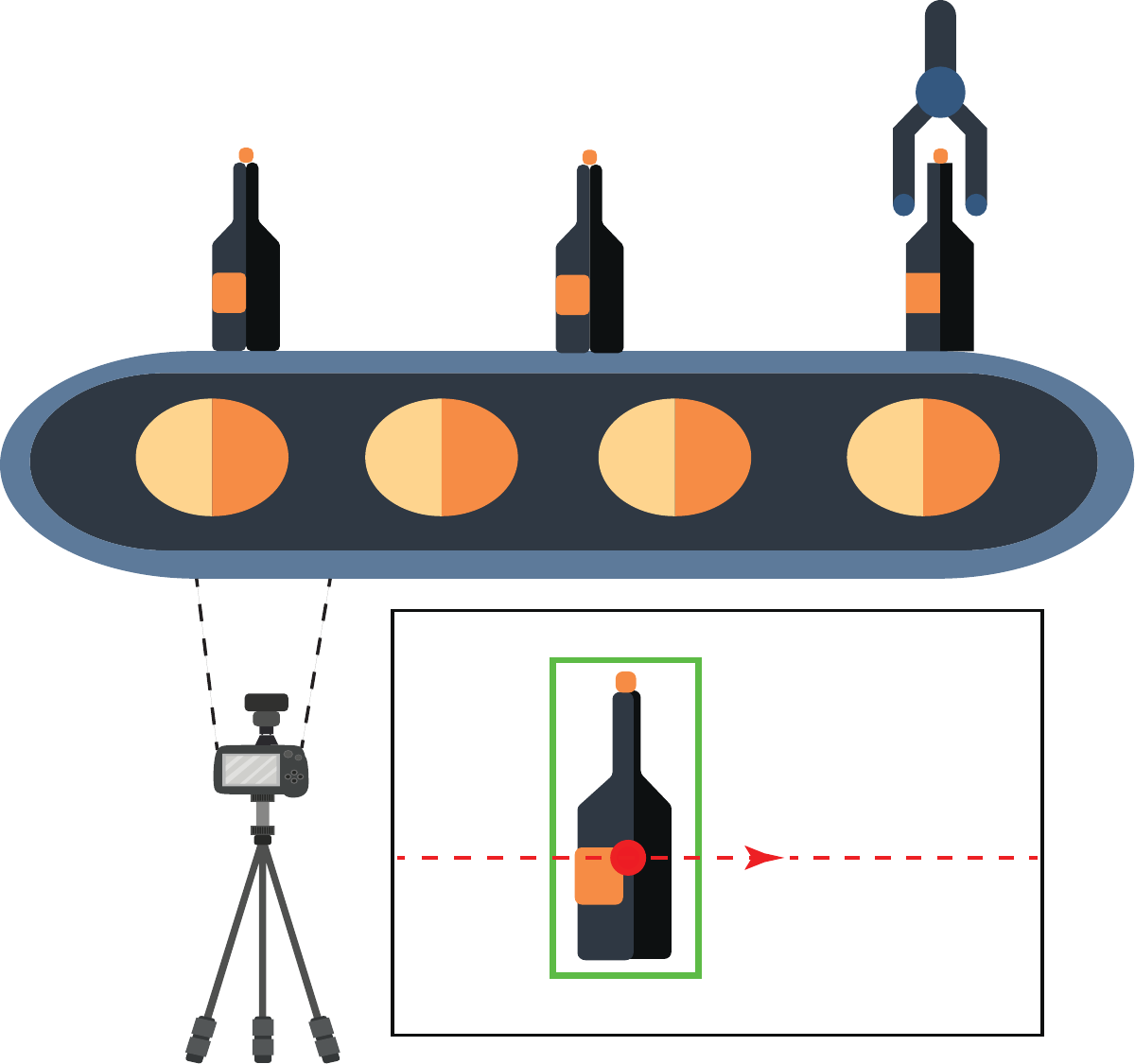}
    \caption{A fixed camera running OD on equally spaced objects on a conveyor belt}
    \label{fig:useCase_prod}
    \end{subfigure}

    \vspace{10pt}

    \begin{subfigure}{.48\textwidth}
    \centering
    \includegraphics[width=0.9\textwidth]{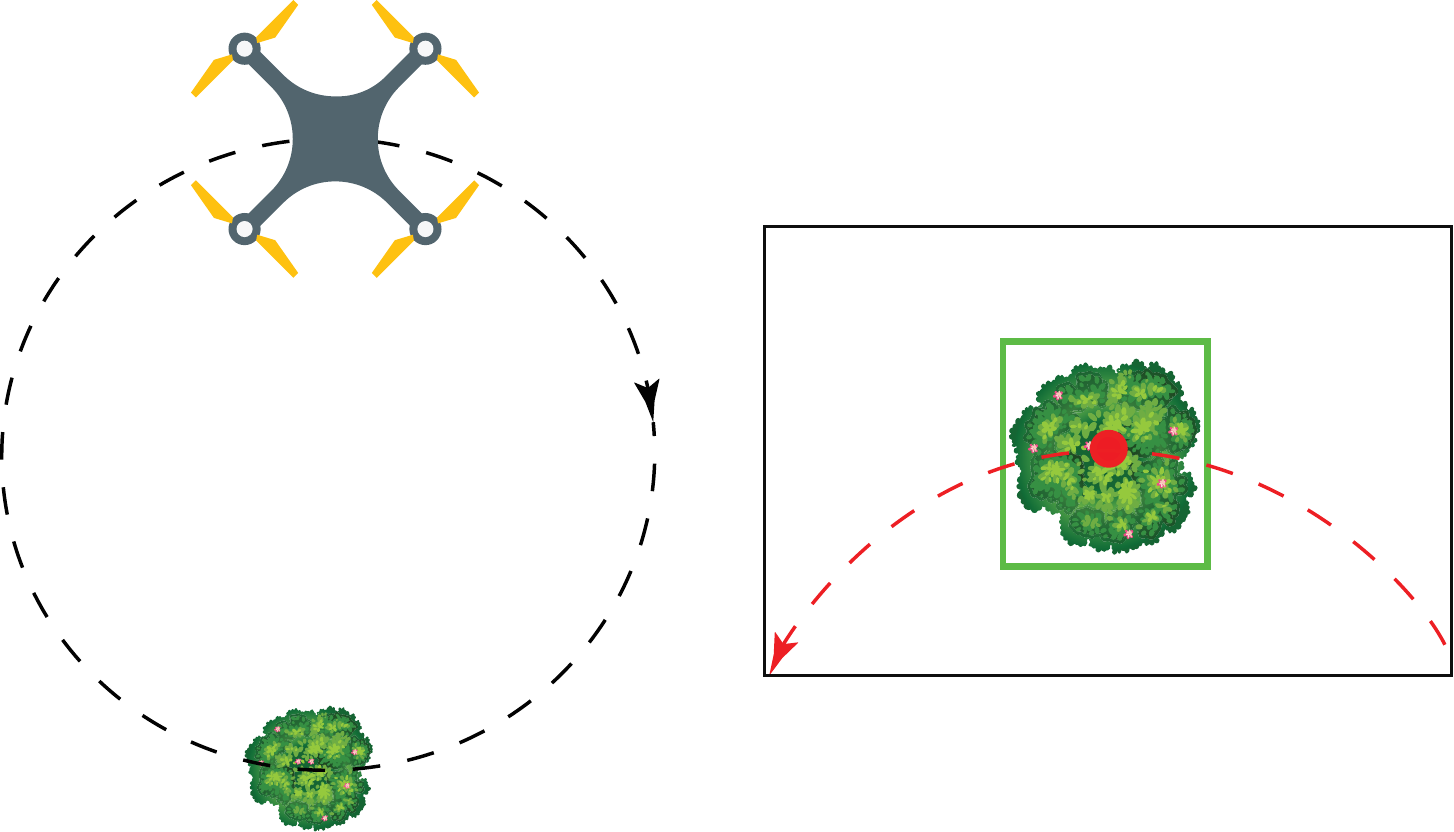}
    \caption{A drone with a periodic motion running OD on fixed objects.}
    \label{fig:useCase_drone}
    \end{subfigure}

    \caption{Examples of use cases for the detection of objects with periodic trajectories.}
    \label{fig:useCase}
\end{figure}

In this paper, we consider the specific case of OD in streams of images, where the objects to be detected are having a periodic motion in the video frames. In order to better understand this setting, two practical use cases are presented in \figurename~\ref{fig:useCase}. In \figurename~\ref{fig:useCase_prod}, an OD model is running on the frames recorded by a fixed camera, to detect objects equally spaced on an automated production line. This scenario has important practical applications in industry~\cite{prodLine1, prodLine2, prodLine3}, where a strong OD model can serve to detect missing parts or to identify regions of interest to conduct more precise quality analysis. On another note, in \figurename~\ref{fig:useCase_drone}, the observed object is fixed but the camera, mounted on a robotic platform, is moving under a periodic motion. Autonomous drones, equipped with cameras and having periodic motions, also have many practical applications such as crops monitoring~\cite{crop_drone}, damage identification in infrastructures~\cite{damage_drone} or intrusion detection~\cite{security_drone}. In these scenarios, robust methods for detection of fixed objects can serve directly to conduct these analyses or as a way to improve localization algorithms. In both cases, the expected BBs appear periodically in the video frame and have the same spatio-temporal trajectories, eventually with some noise.


In previous works, different computer vision problems dealing with objects in motion have been studied. For instance, the similarity between frames has been measured in order to identify and characterize periodic motions~\cite{periodIdentification}. K{\"a}s and Nicolas carried out an approach for trajectory estimation from compressed video files~\cite{compressedTrajEstimation}. A survey about the identification of patterns in spatiotemporal trajectories has been proposed~\cite{patternTraj}, as well a method for pixel-level identification of moving objects in video~\cite{movingOD}. Finally, the periodic motion of the studied objects has been leveraged in order to conduct better anomaly detection~\cite{anomalyPeriodic}. In this paper we aim to leverage the periodic motion to improve the performance of an OD model. To the best of our knowledge, despite its relevance in various practical applications, this problem has never been formally studied.

In the setting of this paper, we consider that an object detection model is already available for the studied objects, but that it is imperfect, i.e. making recognition errors (both false positives and false negatives). Such weak object detection models can occur when few training data are available or when training is performed on 3D CAD models, among other reasons. Under the only assumption that the detected objects have periodic trajectories in the video frames, we aim to propose a post-processing method to filter out the wrong OD predictions. In order to do this, training data are first gathered by running the OD model on the real system and recording the predicted BBs. These spatiotemporal data are then used to train a Gaussian Process~\cite{GP} (GP) to model the periodic trajectories of the BBs. Finally, at inference time, a set of rules on the GP predictions is proposed to discard the wrong predictions of the OD model. We note that in the case where the false positive ratio of the initial OD model is too high, a manual data cleaning of the GP training set must be performed, otherwise an automated outlier detection method, based on DBSCAN clustering~\cite{dbscan}, is proposed in this work.

Our approach is validated on realistic simulated data generated from different periodic trajectories, different levels of initial OD performance, and different levels of noise in the BBs positioning. The results on these datasets show that the proposed GP filtering approach can improve the initial OD model precision-recall curves by a large margin. Besides improving OD robustness, this method also has the advantages of being fast, easy to train and it requires no prior knowledge about the objects motion.


The paper is organized as follows. First, notations are introduced and preliminary concepts necessary to understand the approach and its experimental validation are presented in Section~\ref{sec:notations}. Then, in Section~\ref{sec:method}, both training and inference steps of the proposed GP filtering method are explained in details. The proposed validation on realistic synthetic datasets is presented in Section~\ref{sec:expe}, together with the discussion of the results. Finally, conclusions and directions for future work are exposed in Section~\ref{sec:conclusion}.

\section{Prerequisites and notations}
\label{sec:notations}

In this section, the notations used throughout this paper are introduced and some essential concepts to understand the method are presented. All along this paper, the different steps of the method are illustrated using a simulated toy example.

\subsection{Characterization of an OD model}

Let $\varphi_{\theta}$ be an object detection model, which weights~$\theta$ have already been optimized with respect to the detection task at hand. For a given image $I$, $\varphi_{\theta}(I)$ returns a list of $(l, c, s)$ tuples, where $l$ consists of four coordinates that identify a Bounding Box (BB), $c$ is the predicted category of the object in the BB, and $s$ is the confidence score that the BB contains object~$c$. In order to simplify the notations, we consider that $\varphi_{\theta}$ is trained to recognize a single object. Thus, predictions can be represented by $(l, s)$ tuples only. Extending the method to multiple objects simply consists in independently replicating the approach for each of the predicted classes.

For a specific inference situation, e.g. OD on a given production line, the model performance is fully characterized by its precision-recall (PR) curve. In order to better explain what a PR curve is, we need to introduce some intermediate concepts. First of all, we introduce the confidence threshold $\tau_s$, such that a given prediction $(l, s)$ is kept if $s > \tau_s$ and discarded otherwise. Then, for a given ground-truth BB $l^*$, we define the Intersection over Union (IoU) of $l$ and $l^*$ as the ratio between the intersection area and the union area of the two BBs. After fixing a threshold on IoU ($\tau_{\text{IoU}}$) that complies with the expectations of the application, we can state that $l$ an $l^*$ represent the same object if $\text{IoU}(l, l^*) > \tau_{\text{IoU}}$. From these definitions, we can then say that a detection $(l, s)$ is a true positive (TP) if $s > \tau_s$ and if there exists a ground truth BB $l^*$ such that $\text{IoU}(l, l^*) > \tau_{\text{IoU}}$. If there is not any ground truth satisfying the latter condition, $(l, s)$ is a false positive (FP). Conversely, when a ground truth BB is left without any matching predicted detection, it counts as a false negative (FN).  Finally, in case multiple predictions correspond to the same ground-truth, only the one with the highest confidence score counts as a TP, while the others are considered FP.

With the above definitions, for a given dataset, we can then define precision as the total number of TP divided by the total number of model predictions:
\begin{equation}
Precision = \frac{\sum{TP}}{\sum{TP} + \sum{FP}}.
\end{equation}
A high precision means that most of the predicted BBs had a corresponding ground truth. In other words, the object detector is not producing bad predictions. On the other hand, recall is defined as
\begin{equation}
Recall = \frac{\sum{TP}}{\sum{TP}+\sum{FN}}.
\end{equation}
A high recall means that most of the ground truth BBs had a corresponding prediction. In other words, the object detector finds most objects in the images. Both Recall and Precision vary between 0 and 1, where 1 means perfection.

For a given OD model $\varphi_{\theta}$, the values of Precision and Recall changes for different values of $\tau_s$. Hence, to characterize $\varphi_{\theta}$, we use the Precision-Recall curve, which is plotted by sampling different values of $\tau_s$ and reporting the corresponding PR points in the plot. Besides the model and the value chosen for $\tau_{\text{IoU}}$, the PR curve also depends on the test dataset on which it is computed. For this reason, we talk about the PR curve with respect to a specific inference situation, as the model could perform differently in another environment.

\begin{figure}[!ht]
    \centering
    \includegraphics[width=0.45\textwidth]{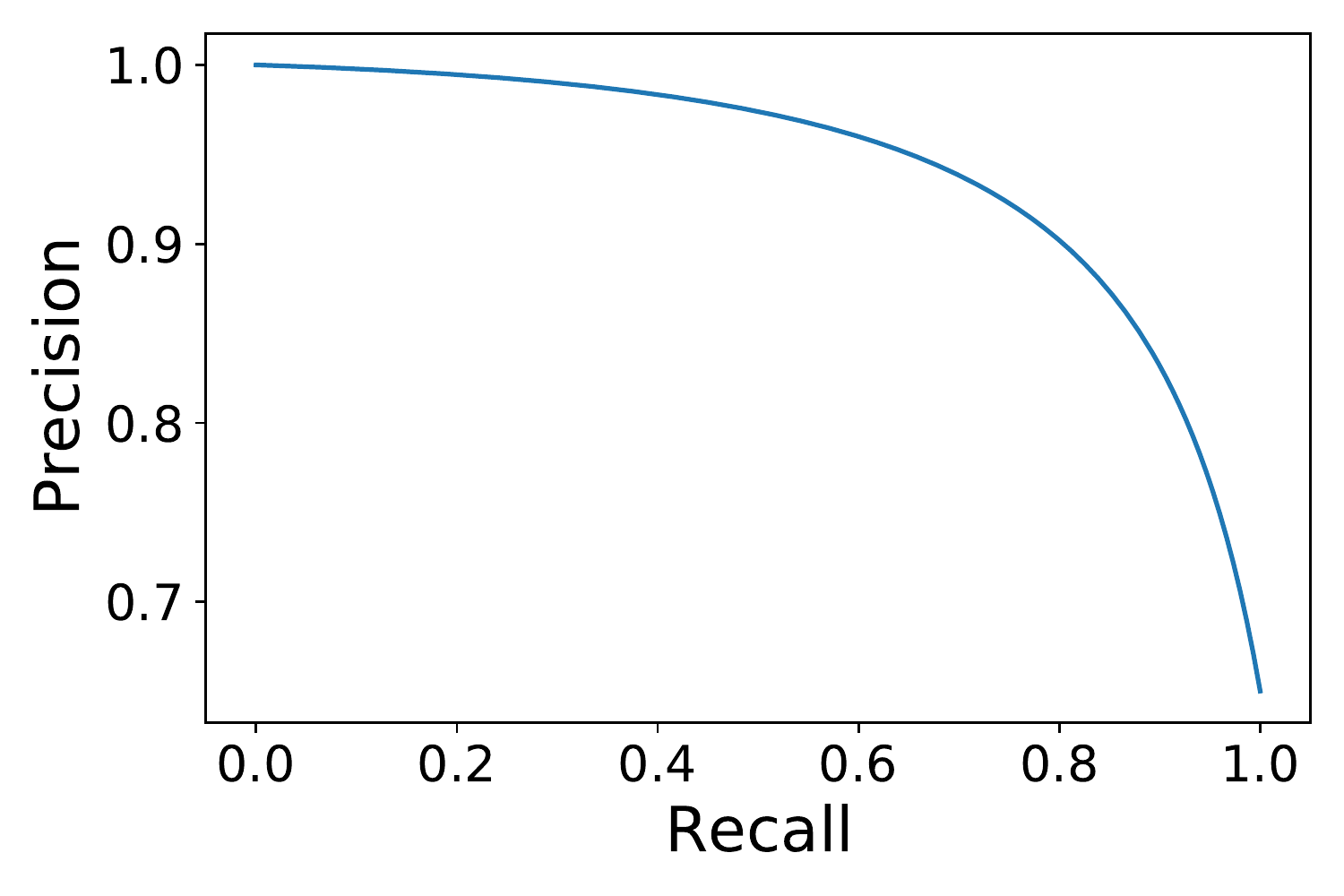}
    \caption{Precision-Recall curve for the toy OD model used to illustrate the method.}
    \label{fig:toy_PR}
\end{figure}

The PR curve of the model used for the simulated toy example is presented in \figurename~\ref{fig:toy_PR}. It can be observed that, for a recall of 0 ($\tau_s = 1$) the precision is 1, which makes sense as when there is no prediction at all there is also no wrong prediction. Then, when $\tau_s$ decreases, the number of predicted BBs increases, thus making the recall increase and the precision decrease.

\subsection{Periodic trajectory of an object}

In the setting studied in this paper, the object that must be detected by $\varphi_{\theta}$ has a periodic motion with respect to the video frames. This periodic trajectory is defined in the three-dimensional space $(x, y, t)$, where $x$ and $y$ are the pixel coordinates in the video frames and $t$ is the time. A spatio-temporal periodic trajectory $(\Gamma, T)$ is represented by both
\begin{itemize}
    \item a period $T$,
    \item a nominal function $\Gamma: t \xrightarrow{} \{x, y\}$, defining the position of the center of the object in the video frame at a given time. $\Gamma$ is defined in $\mathbf{R}^+$, and we have $\Gamma(t + T)~=~\Gamma(t)$. We also note that $\Gamma$ output can be \textit{Null} when the object is not in the video frame.
\end{itemize}

\begin{figure}[!ht]
\centering
    \begin{subfigure}{.48\textwidth}
    \centering
    \includegraphics[width=0.9\textwidth]{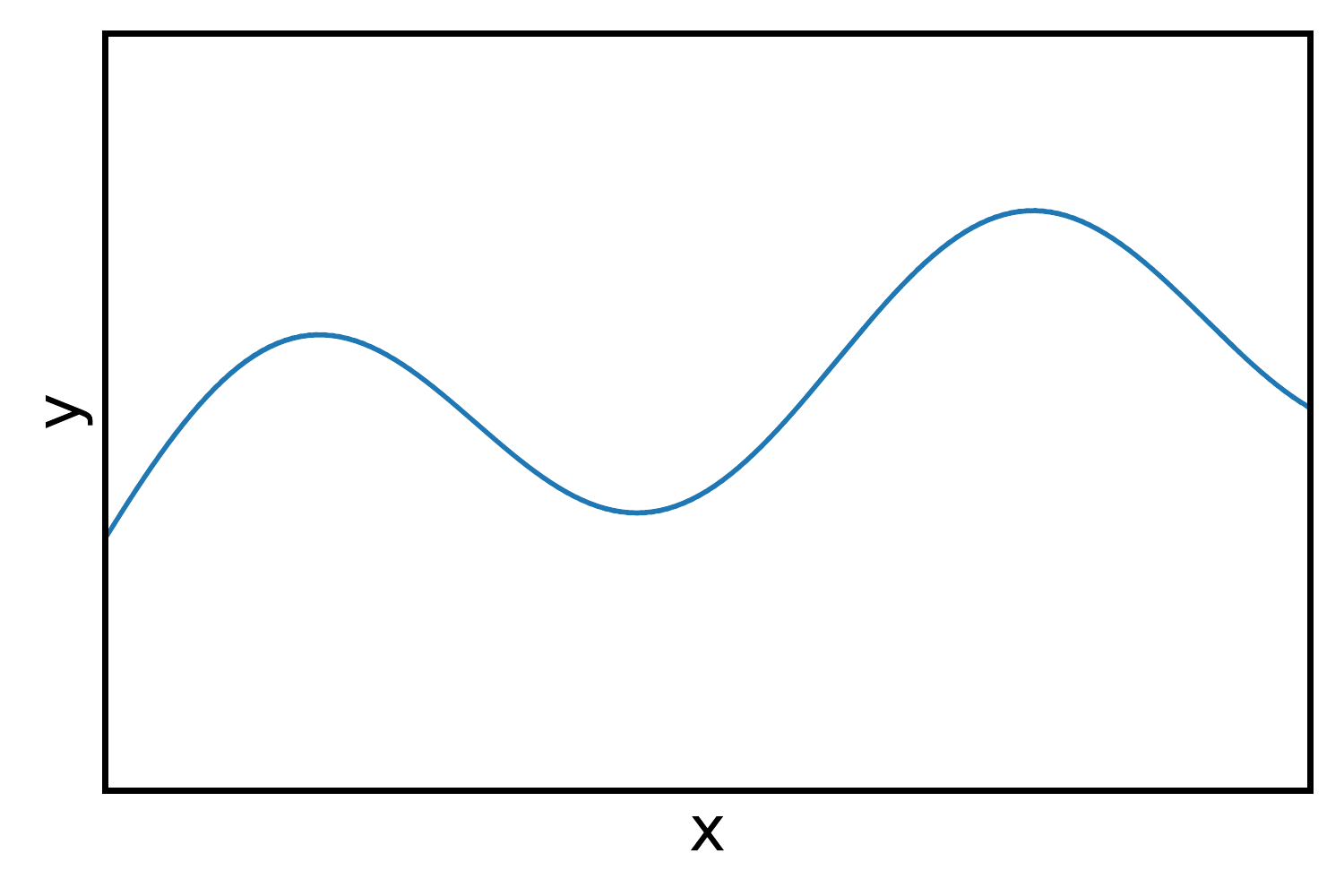}
    \caption{Nominal support line for the center of the object bounding box in the image plane.}
    \label{fig:toyTraj_xy}
    \end{subfigure}

    \begin{subfigure}{.48\textwidth}
    \centering
    \includegraphics[width=0.9\textwidth]{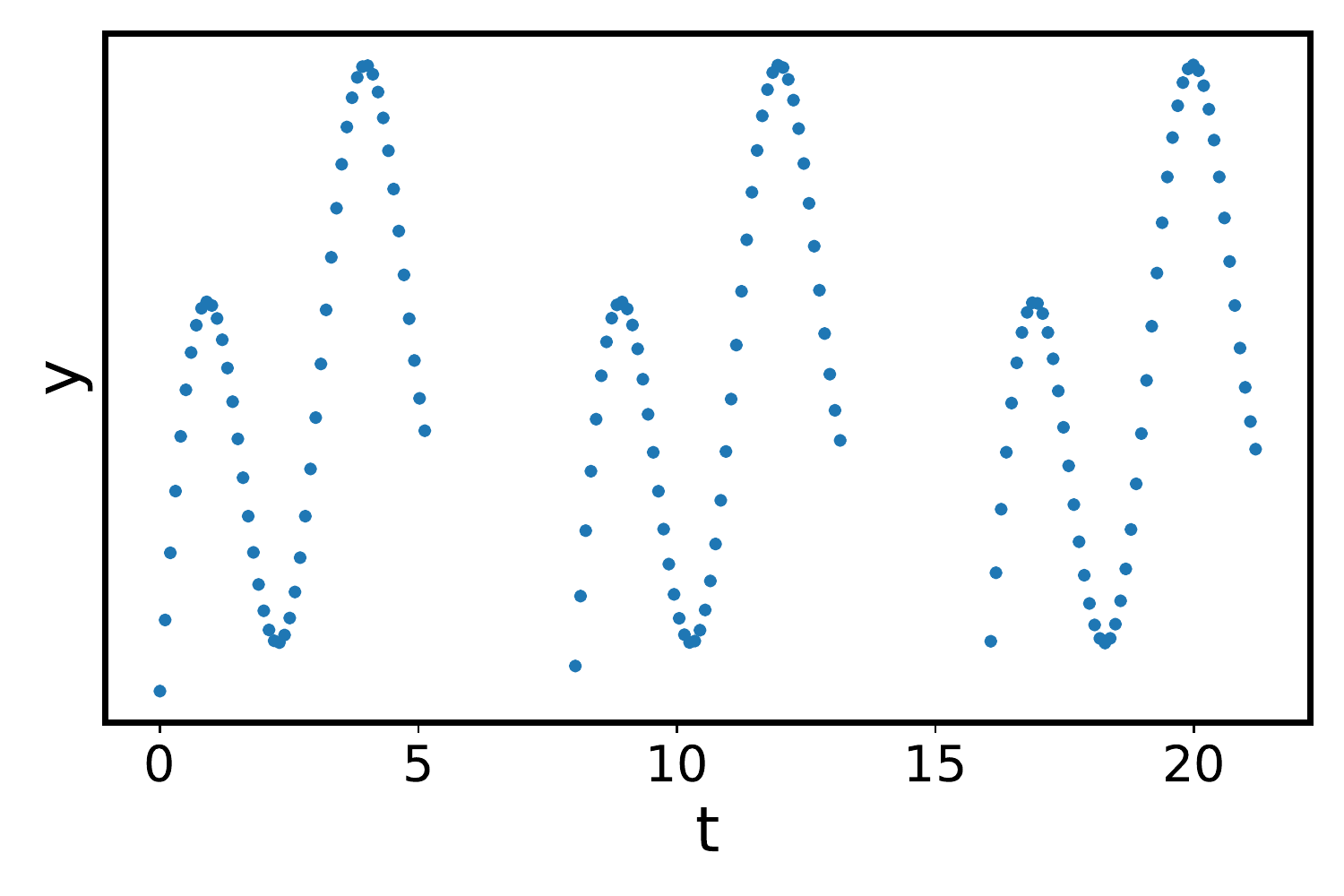}
    \caption{Three periods of the nominal trajectory in the ty-plane.}
    \label{fig:toyTraj_ty}
    \end{subfigure}

    \caption{Different views of the nominal spatio-temporal periodic trajectory of the center of the object for the toy problem used to illustrate the method.}
    \label{fig:toyTraj}
\end{figure}

The periodic trajectory used for the simulated toy example is shown in \figurename~\ref{fig:toyTraj}. When time passes, the object enters the image plane, the center of the BB moves along the line shown in \figurename~\ref{fig:toyTraj_xy} before exiting the video frames, this process is repeated periodically. The period $T$ is the time between two corresponding points in the trajectories (\figurename~\ref{fig:toyTraj_ty}), for this case it is 8 seconds. The t-x view of this toy trajectory is not presented here as it does not provide much additional information. 

\subsection{Generate a simulated periodic OD process}

Once the PR curve of the OD model $\varphi_{\theta}$ and the periodic trajectory of the object $(\Gamma, T)$ have been defined, the entire OD process of the periodic system can be simulated for any fixed value of the recall. In order to do this, we first need to discretize the time such that each time step corresponds to a video frame recording and OD BBs prediction. For each simulated video frame, $(\Gamma, T)$ is used to define the true position of the BB center. Then, for a given recall value, the PR curve can be used to know the corresponding precision and to compute both $P_{FN}$ and $P_{FP}$, respectively the probability to fail detecting the object and the probability to detect a wrong object. For a given total simulation time, these probabilities are defined as
\begin{equation}
    P_{FN} = 1 - \text{recall},
\end{equation}
\begin{equation}
    P_{FP} = \frac{N_{\text{obj}} \times (1 - P_{FN})}{N_{\text{frames}}} \times \frac{1 - \text{precision}}{\text{precision}},
\label{eq:pfp}
\end{equation}
where $N_{\text{frames}}$ is the number of frames in the generated simulation and $N_{\text{obj}}$ is the number of frames in which an object is actually present. The formula in equation~\ref{eq:pfp} is simply the expected proportion of false positives with respect to the total number of simulated frames.

Once $P_{FN}$ and $P_{FP}$ are known, the OD prediction can be easily simulated frame after frame by sampling the true and false BBs. In addition to simulating the imperfect OD model and the physics of the problem (periodic trajectory), we also add noise to both the $(x, y)$ location and the time recording. The former accounts for the errors in the trajectory and the errors in predicted BB positions, while the latter accounts for errors in the time measurement process (e.g. communication lag).

\begin{figure}[!ht]
    \centering
    \includegraphics[width=0.48\textwidth]{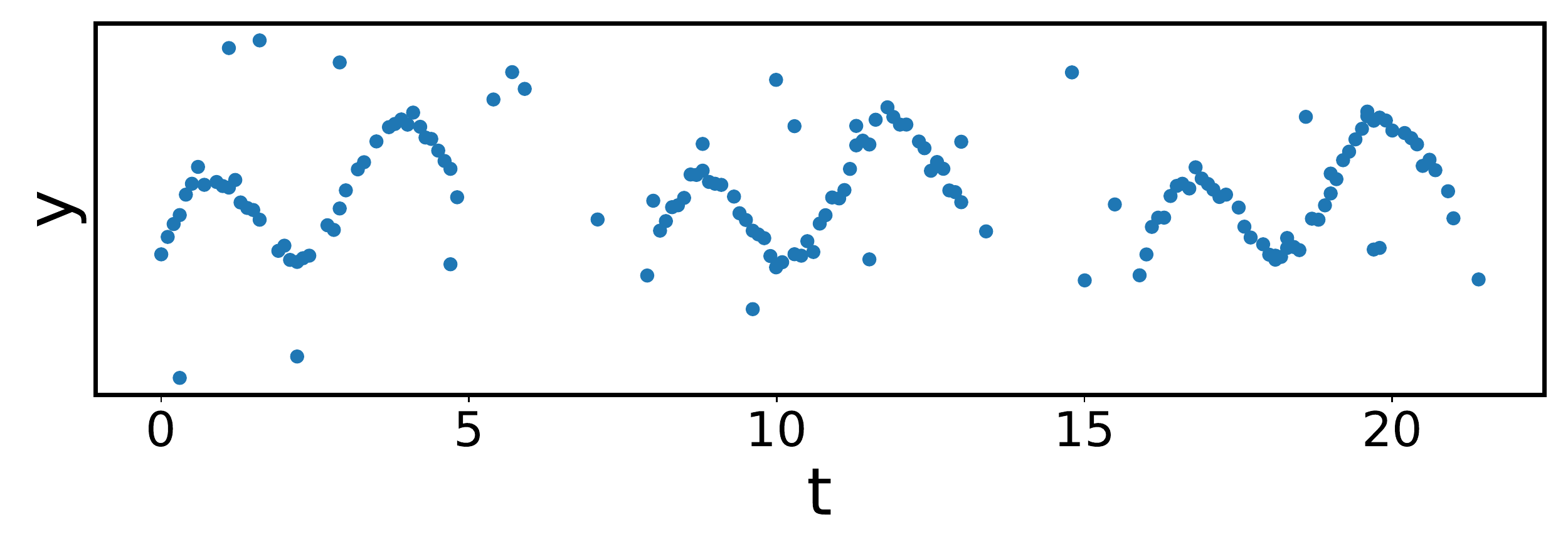}
    \caption{Simulation of 3 periods of the toy example (recall=0.9).}
    \label{fig:toy_noisy}
\end{figure}

In practice, for our toy example a new frame is simulated every 0.1 second (sampling time). We use Gaussian noise with mean $0$ and standard deviation of $0.1$ for position (diagonal covariance matrix) and $5\times10^{-3}$ for time. Three periods of the noisy simulated OD data for the toy problem can be seen in \figurename~\ref{fig:toy_noisy}.

\section{Proposed approach}
\label{sec:method}

\begin{figure*}
\centering

\begin{subfigure}{\textwidth}
\centering
\begin{tikzpicture}

\node[inner sep=0pt] (system) at (0,0)
{\includegraphics[width=.15\textwidth]{useCase_prod.pdf}};
\node[inner sep=0pt, font = \bf] (system_txt_down) at ($(system.center)+(0,1.6)$) {\small OD system};
\node[inner sep=0pt, font = \bf] (system_txt_up) at ($(system_txt_down.north)+(0,0.25)$) {\small Real periodic};
\node[inner sep=0pt] (text_train_down) at ($(system.center)+(2.5,0.3)$) {\small training data};
\node[inner sep=0pt] (text_train_up) at ($(text_train_down.north)+(0,0.25)$) {\small Collect};

\node[inner sep=0pt] (noisy) at ($(system.center)+(5,0)$)
{\includegraphics[width=.15\textwidth]{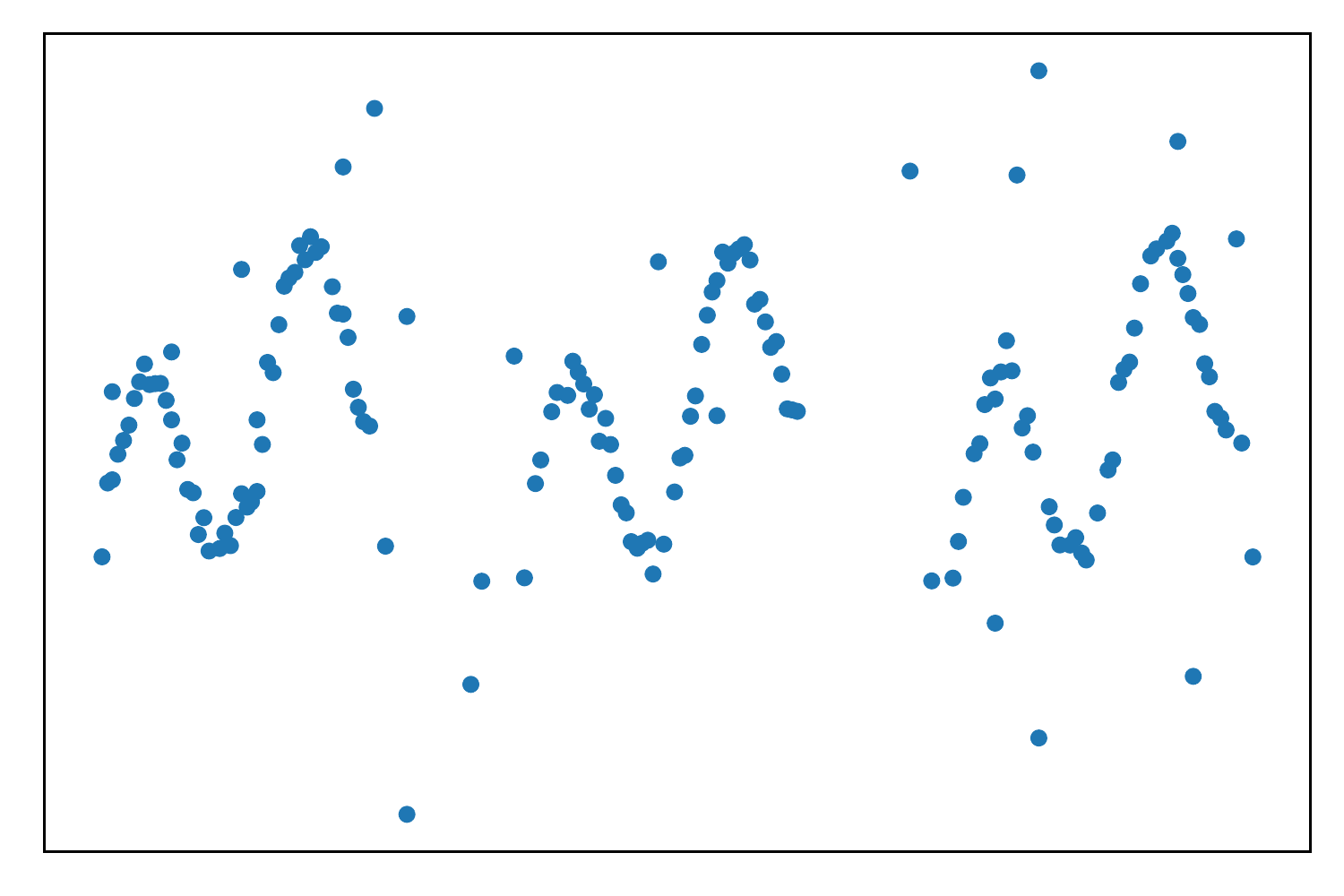}};
\node[inner sep=0pt, font = \bf] (noisy_txt_down) at ($(noisy.center)+(0,1.6)$) {\small training data};
\node[inner sep=0pt, font = \bf] (noisy_txt_up) at ($(noisy_txt_down.north)+(0,0.25)$) {\small Raw noisy};
\node[inner sep=0pt, font = \bf] (noisy_x) at ($(noisy.south)+(0,-0.1)$) {t};
\node[inner sep=0pt, font = \bf] (noisy_y) at ($(noisy.west)+(-0.05,0)$) {y};
\node[inner sep=0pt] (text_clean_down) at ($(noisy.center)+(2.5,0.3)$) {\small clean data};
\node[inner sep=0pt] (text_clean_up) at ($(text_clean_down.north)+(0,0.25)$) {\small Align \&};

\node[inner sep=0pt] (align) at ($(noisy.center)+(5,0)$)
{\includegraphics[width=.15\textwidth]{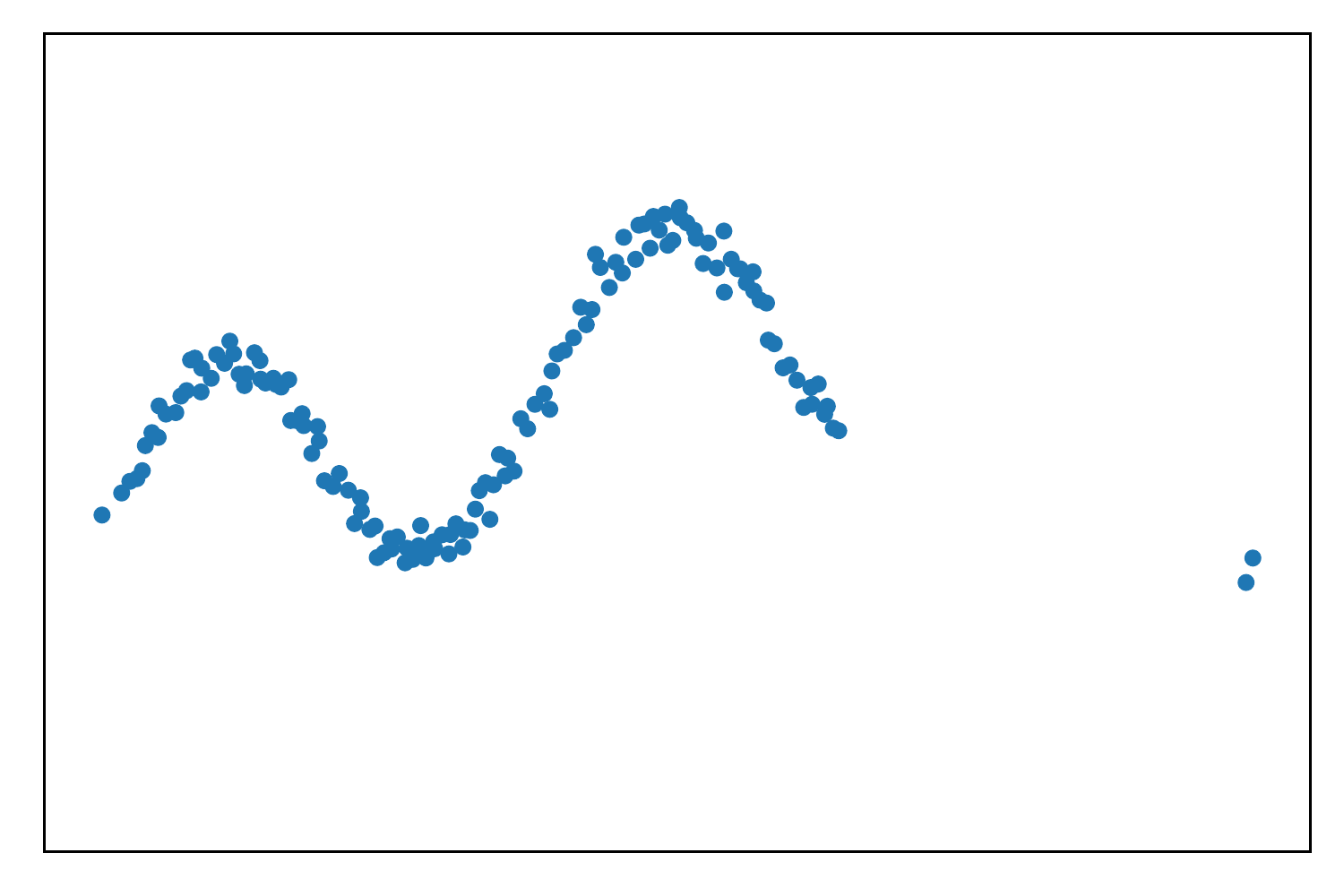}};
\node[inner sep=0pt, font = \bf] (align_txt_down) at ($(align.center)+(0,1.6)$) {\small training data};
\node[inner sep=0pt, font = \bf] (align_txt_up) at ($(align_txt_down.north)+(0,0.25)$) {\small Clean aligned};
\node[inner sep=0pt, font = \bf] (align_x) at ($(align.south)+(0,-0.1)$) {t mod $\widehat{\text{T}}$};
\node[inner sep=0pt, font = \bf] (align_y) at ($(align.west)+(-0.05,0)$) {y};
\node[inner sep=0pt] (text_gp_down) at ($(align.center)+(2.5,0.3)$) {\small GP model};
\node[inner sep=0pt] (text_gp_up) at ($(text_gp_down.north)+(0,0.25)$) {\small Train};

\node[inner sep=0pt] (trained) at ($(align.center)+(5,0)$)
{\includegraphics[width=.15\textwidth]{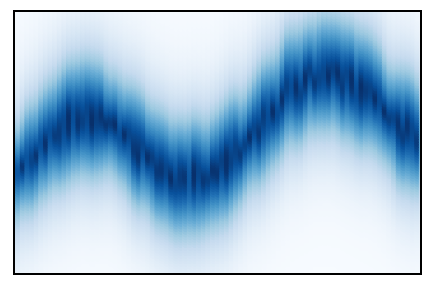}};
\node[inner sep=0pt, font = \bf] (trained_txt_up) at ($(trained.center)+(0,1.6)$) {\small predictive model};
\node[inner sep=0pt] (trained_txt_down) at ($(trained_txt_up.south)+(0,-0.25)$) {\small $\text{GP}(x, y) = \hat{t} \pm \hat{\sigma}$};

\node[inner sep=0pt, font = \bf] (trained_txt_up2) at ($(trained_txt_up.north)+(0,0.25)$) {\small Trained GP trajectory};
\node[inner sep=0pt, font = \bf] (trained_x) at ($(trained.south)+(0,-0.1)$) {x};
\node[inner sep=0pt, font = \bf] (trained_y) at ($(trained.west)+(-0.05,0)$) {y};

\draw[->, line width = 1] ($(system.east)+(0.02,0)$) -- ($(noisy_y.west)+(-0.02,0)$);
\draw[->, line width = 1] ($(noisy.east)+(0.02,0)$) -- ($(align_y.west)+(-0.02,0)$);
\draw[->, line width = 1] ($(align.east)+(0.02,0)$) -- ($(trained_y.west)+(-0.02,0)$);

\end{tikzpicture}

\caption{Training}
\label{fig:method_training}
\end{subfigure}

\vspace{10pt}

\begin{subfigure}{\textwidth}
\centering

\begin{tikzpicture}

\node[inner sep=0pt] (frame) at (0,0)
{\includegraphics[width=.12\textwidth]{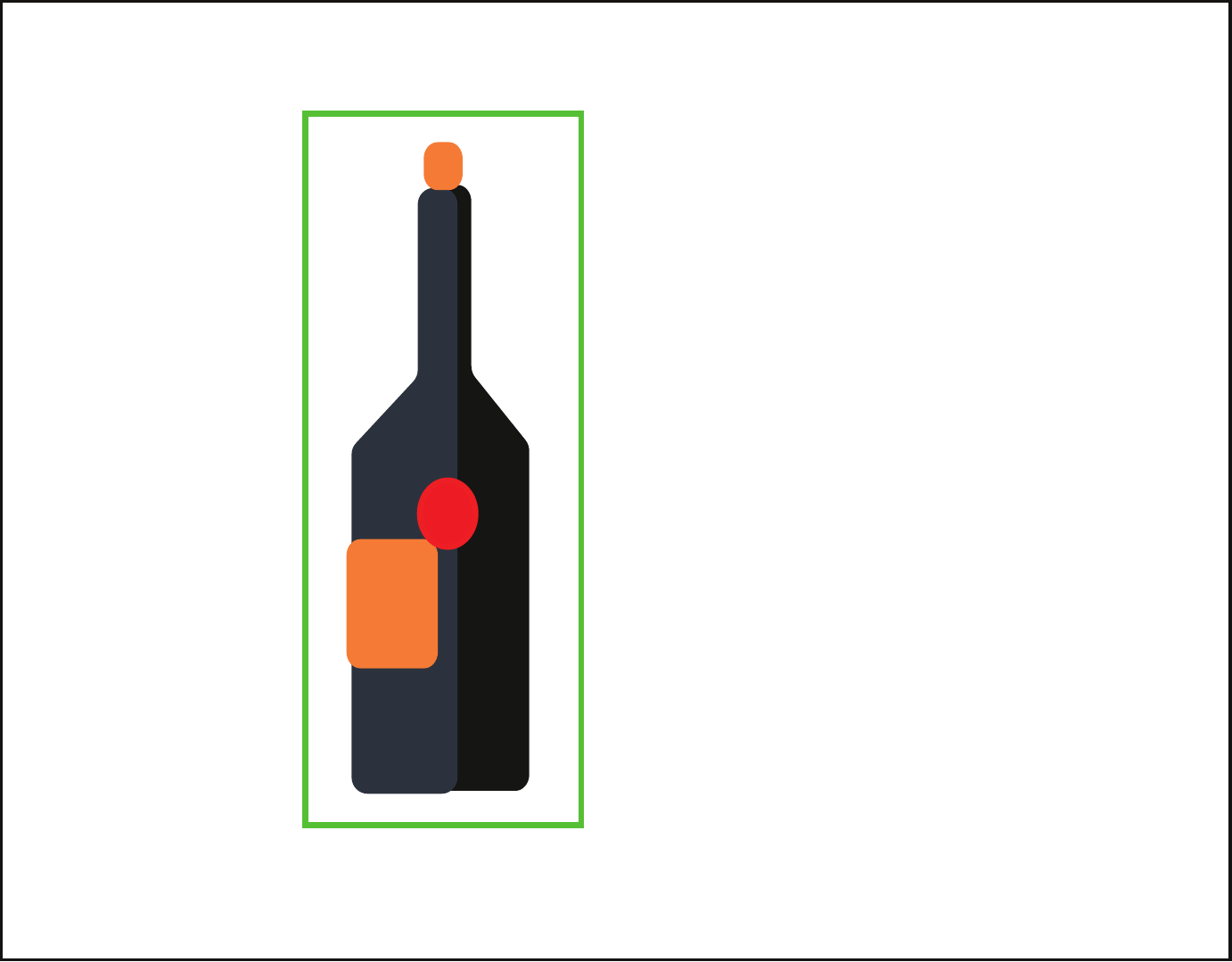}};
\node[inner sep=0pt, font = \bf] (frame_txt) at ($(frame.center)+(0,1.5)$) {\small New frame};
\node[inner sep=0pt] (text_frame_down) at ($(frame.center)+(2,0.3)$) {\small OD model};
\node[inner sep=0pt] (text_frame_up) at ($(text_frame_down.north)+(0,0.25)$) {\small Running};

\node[inner sep=0pt] (center) at ($(frame.center)+(3.8,0)$)
{\huge $\{x, y\}$};
\node[inner sep=0pt, font = \bf] (center_txt_up) at ($(center.center)+(0,1.5)$) {\small Bounding box};
\node[inner sep=0pt, font = \bf] (center_txt_down) at ($(center_txt_up.south)+(0,-0.15)$) {\small center};

\node[inner sep=0pt] (text_pred_down) at ($(center.center)+(1.7,0.3)$) {\small GP model};
\node[inner sep=0pt] (text_pred_up) at ($(text_pred_down.north)+(0,0.25)$) {\small Running};

\node[inner sep=0pt] (gp) at ($(center.center)+(3.4,0)$)
{\huge $\{\hat{t}, \hat{\sigma}\}$};
\node[inner sep=0pt, font = \bf] (gp_txt_up) at ($(gp.center)+(0,1.5)$) {\small Predicted time};
\node[inner sep=0pt, font = \bf] (gp_txt_down) at ($(gp_txt_up.south)+(0,-0.25)$) {\small \& uncertainty};

\node[inner sep=0pt] (cond1) at ($(gp.center)+(4,0.4)$)
{\huge $\hat{\sigma} < \sigma_{\text{max}}$};
\node[inner sep=0pt] (cond2) at ($(gp.center)+(4,-0.4)$)
{\huge $|\hat{t} - t'| < \sigma_{\text{max}}$};
\node[inner sep=0pt, font = \bf] (cond_txt_down) at ($(cond1.center)+(0,1.1)$) {\small Filtering conditions};

\node[inner sep=0pt, anchor= west] (true) at ($(cond1.center)+(3.6,0.5)$)
{\large Keep};
\node[inner sep=0pt, anchor=west] (false) at ($(cond2.center)+(3.6,-0.5)$)
{\large Discard};

\draw [] ($(cond2.west)+(-0.1,1.4)$) rectangle ($(cond2.east)+(0.1,-0.6)$);

\draw[->, line width = 1] ($(frame.east)+(0.02,0)$) -- ($(center.west)+(-0.02,0)$);
\draw[->, line width = 1] ($(center.east)+(0.02,0)$) -- ($(gp.west)+(-0.02,0)$);
\draw[->, line width = 1] ($(gp.east)+(0.02,0)$) -- ($(cond2.west)+(-0.12,0.4)$);
\draw[->, line width = 1] ($(cond2.east)+(0.12,0.4)$) -- ($(true.west)+(-0.02,0)$) node [midway, above, sloped] (true_txt) {\small True};
\draw[->, line width = 1] ($(cond2.east)+(0.12,0.4)$) -- ($(false.west)+(-0.02,0)$) node [midway, below, sloped] (TextNode) {\small False};

\end{tikzpicture}
\caption{Inference}
\label{fig:method_inference}
\end{subfigure}

\caption{Overview of the proposed filtering method for robust detection of objects under periodic motion.}
\label{fig:method}
\end{figure*}

An overview of the approach proposed in this paper can be seen in \figurename~\ref{fig:method}. It consists in gathering training data by running the real world OD system for some periods and recording the center of the predicted BBs as well as their associated time. Then, these noisy data are pre-processed in order to align the trajectories and filter the false positives. The clean data are then used to fit a Gaussian Process model which predicts the time from the xy coordinates of a BB center. At inference, when a new BB is predicted by the OD model, the trained GP model is used to filter out wrong boxes in real time. In this section, the different steps of the proposed periodic OD filtering approach are presented in details. The toy example presented in Section~\ref{sec:notations} is used to illustrate these steps.

\subsection{Gaussian Process training}

First, the methodology to fit a GP model to the periodic trajectory is presented in the following steps.

\subsubsection{Collecting training data}

The collection of training data consists in running the real OD system for a number of periods $N$, defined by the user. To run the real OD system, a point in the PR curve of the OD model $\varphi_{\theta}$ should be selected by fixing the value of $\tau_s$. A configuration with high recall should be selected in order to get sufficient TPs for the GP training. In practice, the best results were obtained for recall$=0.9$, and we found that $N=5$ is sufficient to train a good filtering model. Then, for each prediction made by $\varphi_{\theta}$, the BB center as well as the time since the beginning of the experiment are recorded. This generates a set of three-dimensional data points $(x, y, t)$. \figurename~\ref{fig:toy_noisy} shows the projection on the ty-plane of this dataset for the toy example with $N = 3$.

\subsubsection{Aligning training data}

\begin{figure}[!ht]
\centering
    \begin{subfigure}{.48\textwidth}
    \centering
    \includegraphics[width=0.9\textwidth]{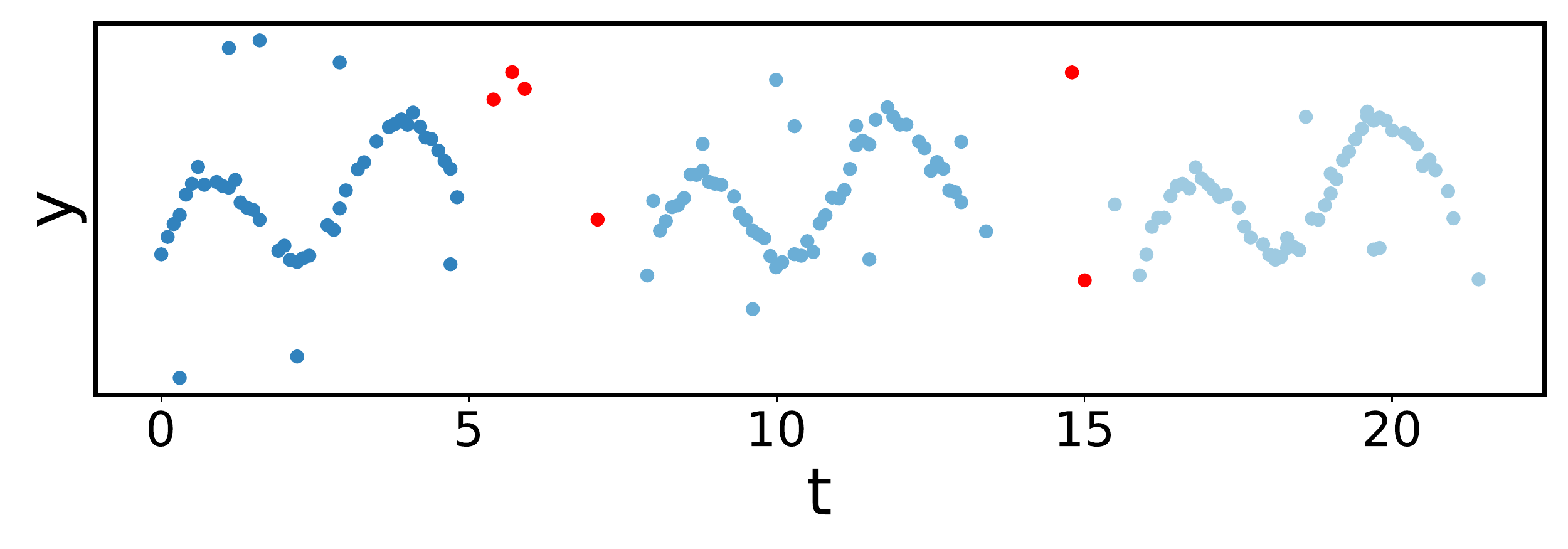}
    \caption{DBSCAN clustering results. Different shades of blue represent different clusters, red points do not belong to any cluster.}
    \label{fig:toyalign_cluster}
    \end{subfigure}

    \begin{subfigure}{.48\textwidth}
    \centering
    \includegraphics[width=0.9\textwidth]{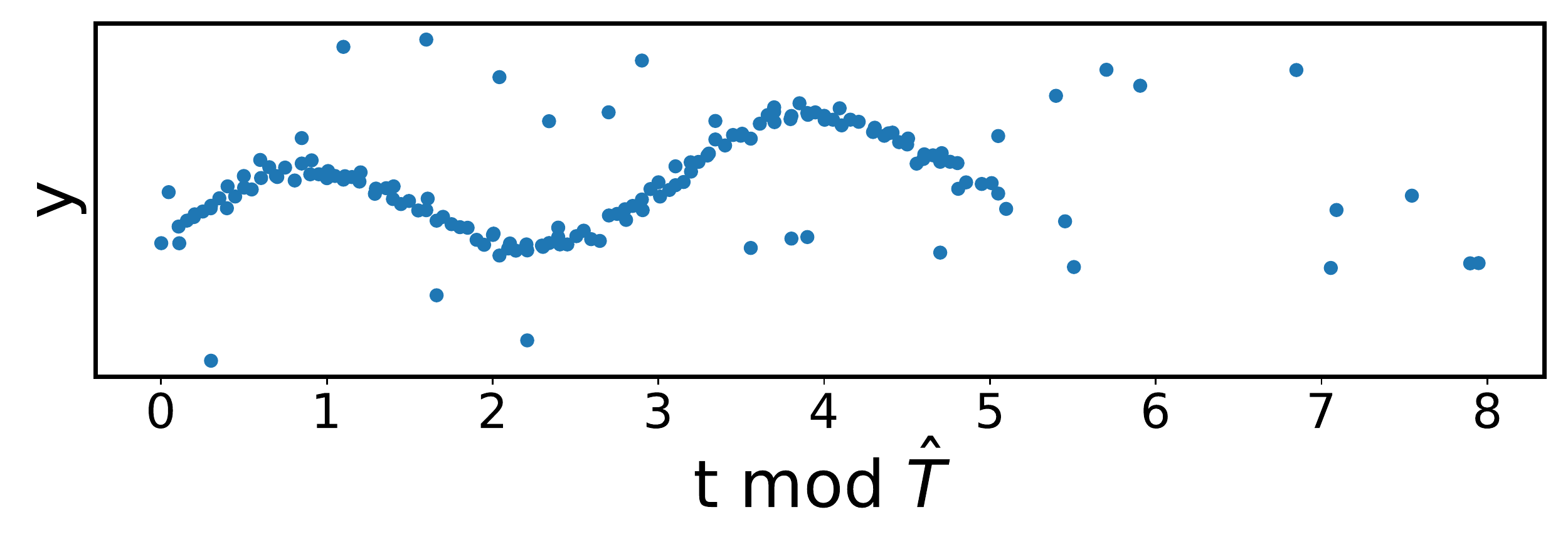}
    \caption{Aligned data, after estimation of period $\widehat{T}$.}
    \label{fig:toyalign_align}
    \end{subfigure}

    \caption{Alignment of raw noisy training data, collected from the periodic object detection toy example.}
    \label{fig:toyalign}
\end{figure}

In order to align training data, the period $T$ of the object motion needs to be identified. To do this, we leverage the fact that the density of points is lower when the object is not actually present in the frame. Indeed, between two object appearances in the video, the predictions only consist of FP, whereas when the object is present there are both TP and FP. Hence, in order to identify the period $T$, we run a density-based clustering algorithm (DBSCAN~\cite{dbscan}) on the one-dimensional projection of the training data along the t-axis. In practice, we use the scikit-learn~\cite{scikit-learn} implementation of DBSCAN with eps = 0.5, min\_samples = 5 and euclidean distances. \figurename~\ref{fig:toyalign_cluster} shows the DBSCAN clusters assignment for the training set gathered from the toy example, with $N = 3$. Once the clusters representing different trajectory instances have been identified, the estimated period $\widehat{T}$ can be computed as follows:
\begin{equation}
\widehat{T} = \text{Med}\left(\left\{t_{\text{min}}^{i+1} - t_{\text{min}}^i \mid i \in |\mathcal{C}| \right\} \cup \left\{t_{\text{max}}^{i+1} - t_{\text{max}}^i \mid i \in |\mathcal{C}|\right\}\right),
\label{eq:T_hat}
\end{equation}
where $Med$ is the median, $|\mathcal{C}|$ is the total number of defined clusters, and $t_{\text{min}}^i$ (respectively $t_{\text{max}}^i$) is the smallest (respectively larger) time value of cluster $i$. Eq.(\ref{eq:T_hat}) simply estimates the distance between corresponding points in two consecutive clusters. Considering both first and last points, as well as using the median, reduce the potential risk of error from including a FP at the extremity of one cluster. Once the period has been estimated from the noisy data, they can be aligned using the \textit{modulo} operation, by replacing the $t$ component by $t~\text{mod}~\widehat{T}$ (\figurename~\ref{fig:toyalign_align}).

\subsubsection{Cleaning training data}

\begin{figure}[!ht]
\centering
    \begin{subfigure}{.48\textwidth}
    \centering
    \includegraphics[width=0.9\textwidth]{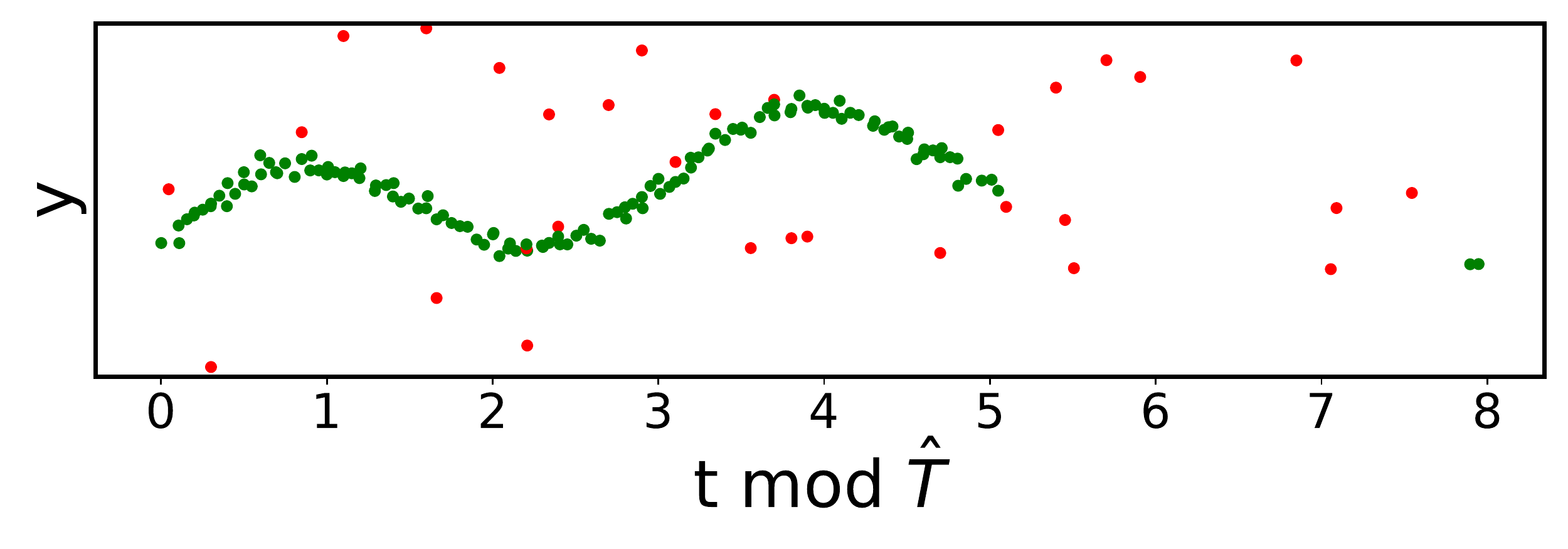}
    \caption{DBSCAN clustering results on (x, y, t mod $\widehat{T}$) data.}
    \label{fig:toyclean_cluster}
    \end{subfigure}

    \begin{subfigure}{.48\textwidth}
    \centering
    \includegraphics[width=0.9\textwidth]{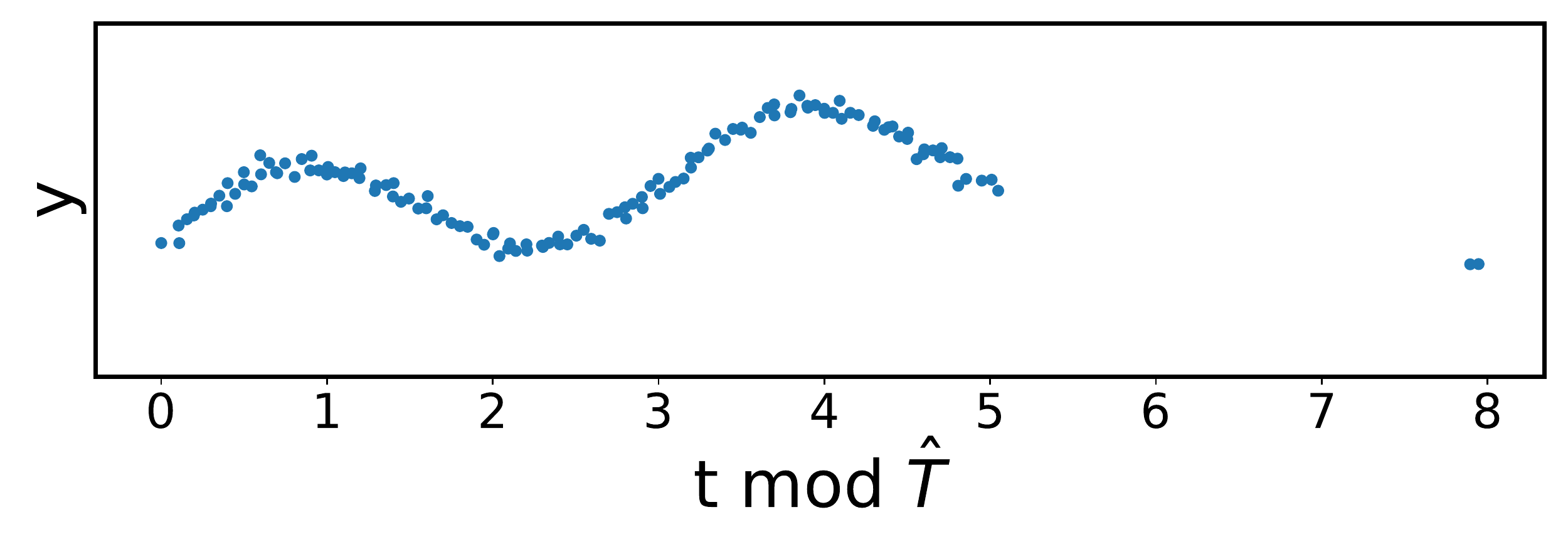}
    \caption{Clean aligned data for GP training, after removal of FPs.}
    \label{fig:toyclean_remove}
    \end{subfigure}

    \caption{Cleaning of noisy aligned training data for the toy example. Identification and removal of false detections.}
    \label{fig:toyclean}
\end{figure}

To remove the outliers (FP) from the aligned training set, we rely on the fact that the density of points should be higher where objects were actually present. In fact, after alignment, there should be around $N$ points (maybe less in case of FN) in the neighborhoods of true predictions, whereas there is no reason for such pattern for FPs. Hence, a second stage of DBSCAN clustering is applied to the $(x, y, t~\text{mod}~\widehat{T})$ 3-dimensional data. The optimal DBSCAN hyper-parameters for this step depend on the number of training period $N$ (estimated from the data) and the noise in the point locations (unknown). Hence, we set the number of neighbors (min\_samples) to $0.8 \times N$ and the optimal eps is computed using the elbow method~\cite{elbow}. \figurename~\ref{fig:toyclean_cluster} shows the results from this clustering on the toy example ($N=3$), all point belonging to a cluster are displayed in green while others are in red. The clean aligned training data obtained after removing the outliers can be seen in \figurename~\ref{fig:toyclean_remove}.

\subsubsection{Gaussian Process training}

\begin{figure}
    \centering
    \includegraphics[width=0.45\textwidth]{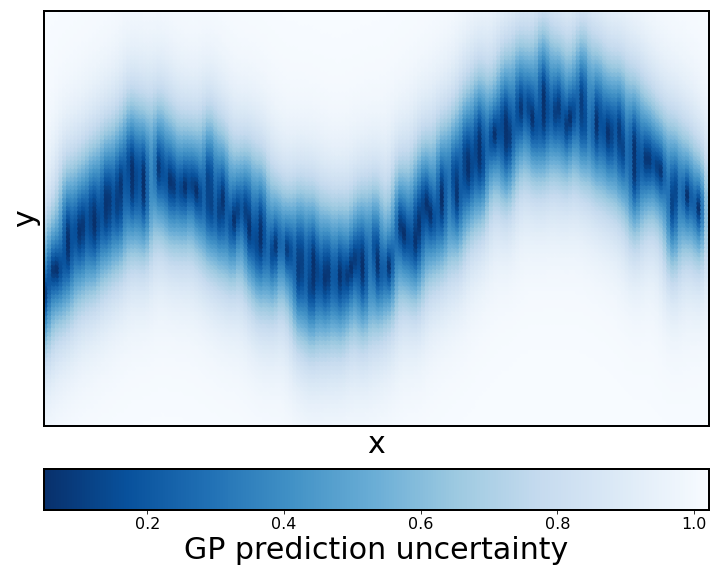}
    \caption{GP prediction uncertainty for every pixel of the video frame for the toy example.}
    \label{fig:toy_gp_uncertainty}
\end{figure}

The clean aligned training set presented above is then used to fit a GP model $\mathcal{GP}$ to the object spatio-temporal trajectory. A GP is a non-parametric Bayesian regression approach that finds a distribution over the possible functions that are consistent with the observed data. For a detailed overview of Gaussian processes, we refer the reader to~\cite{GP}. In order to be useful for BB filtering, $\mathcal{GP}$ is fit to model the following function:
\begin{equation}
    \begin{array}{rccc} f: & I & \to & [0, \widehat{T}], \\
    & (x, y) & \to & t. \end{array}
\end{equation}
In other words, we want to train a predictive model such that, for every pixel in the image, we can predict the trajectory time at which the center of the object is supposed to be at this location. The main advantage of using a GP model is that, for a given input point $(x, y)$, it does not only return the predicted value $\hat{t}$ but also the standard deviation of the model at that precise input point $\hat{\sigma}$. The latter can serve as a proxy for the uncertainty of the model at a given input pixel, it depends on the distance from the closest points seen by the model during training. Having this information is very useful to define a BB filter directly from data, as shown latter in Section~\ref{sec:method_inference}. \figurename~\ref{fig:toy_gp_uncertainty} shows the uncertainty of the $\mathcal{GP}$ predictions for every pixel of the video frame, after training $\mathcal{GP}$ on the training data from \figurename~\ref{fig:toyclean_remove}. We can see that along the object trajectory, $\mathcal{GP}$ is highly confident about its prediction, and the uncertainty increases as we move away from the nominal trajectory line.

\subsubsection{Manual data cleaning for bad initial OD model}

As shown later in the experimental results (Section~\ref{sec:expe}), when the initial OD model $\varphi_{\theta}$ is bad, the approach presented to align and clean training data does not work well. This is because there are too many outliers between trajectories and the clustering fails to identify the period $T$. When this happens, one can fix this problem by manually removing the FPs in the training data before running the data alignment step. After this initial manual cleaning, the approach is the same. Both automatic and manual data cleaning are evaluated in our experiments for different initial OD models.

\subsection{Online bounding box filtering at inference time}
\label{sec:method_inference}
After $\mathcal{GP}$ has been trained by following the approach presented above, it can be used to improve $\varphi_{\theta}$ robustness on the periodic OD task at hand. This is done by filtering BBs online as soon as they appear, in real time. In order to do this, we first need to compute the maximum of uncertainty across all training samples $\sigma_{\text{max}}$. Then, when a new frame is recorded, we run $\varphi_{\theta}$ to obtain BBs predictions. If a BB passes the thresholding test on confidence scores ($s > \tau_s$), its center $(x, y)$ is given as input to $\mathcal{GP}$ for time prediction:
\begin{equation}
    \{\hat{t}, \hat{\sigma}\} = \mathcal{GP}(x, y).
\end{equation}
Then, two filtering tests are computed on the predictions:
\begin{equation}
    \hat{\sigma} < \sigma_{\text{max}},
\label{eq:cond1}
\end{equation}
\begin{equation}
    |\hat{t} - t'| < \sigma_{\text{max}},
\label{eq:cond2}
\end{equation}
where $t'$ is the cyclic execution time ($t' = t~mod~\widehat{T}$, with $t$ being the time since the beginning of the execution).

If Eq.(\ref{eq:cond1}) is true, it means that the point is close to the nominal trajectory of the object. If Eq.(\ref{eq:cond2}) is true, it means that the predicted time corresponds to the actual time of the experiment. Hence, when both conditions are true, it means that the predicted BB is at the right location and at the right time, where the object is supposed to be. In this way, the GP filtering step consists in running both tests on every predicted BB and discarding any prediction that fails at least one of them. We also underline that when the starting time for inference does not match the starting time for the training sample trajectories, the first few predictions verifying condition of Eq.(\ref{eq:cond1}) can be used to match prediction time and run time.

\begin{figure}[!ht]
    \centering
    \includegraphics[width=0.45\textwidth]{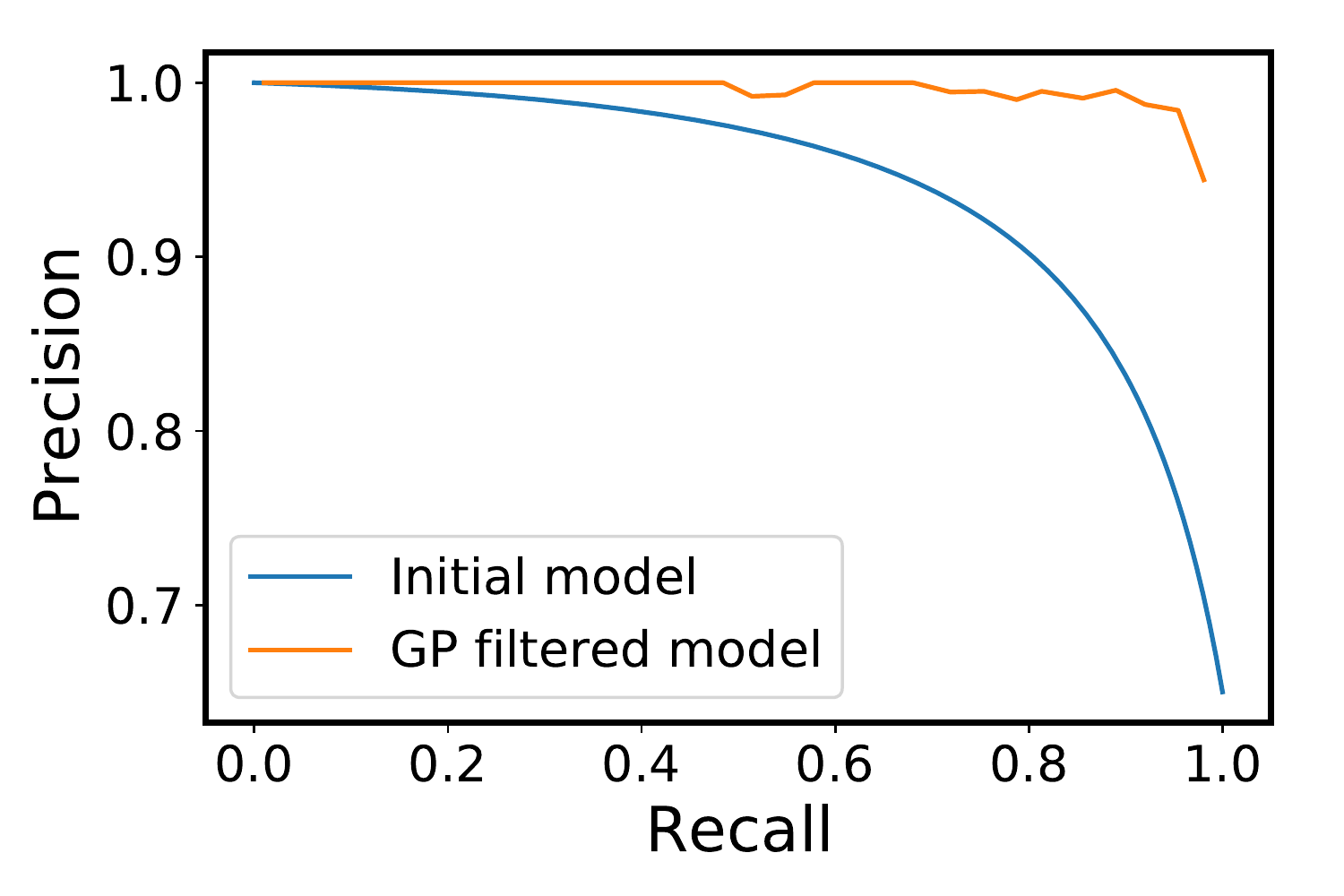}
    \caption{The new PR curve after filtering for the toy problem.}
    \label{fig:toy_newPR}
\end{figure}

After applying this method, some bounding boxes are removed by the GP filtering. On the one hand, if the training went well, most of the boxes discarded should be FPs and, for a given recall value of the original $\varphi_{\theta}$ model, the post-filtering precision should be higher than the original precision. On the other hand, as some TPs might also be removed while no FNs are removed, the post-filtering recall is lower or equal than the original recall. However, if, as expected, the filter removes significantly more FPs than TPs, the confidence threshold $\tau_s$ can be lowered in order to increase the initial recall without lowering the precision, thanks to the filtering. To evaluate the impact of the GP filtering approach, we can run the inference process presented here for all possible values of the initial $\varphi_{\theta}$ recall. Then, by computing the post-filtering recall and precision for each of these values, we can compute the post-filtering PR curve. The new PR curve obtained after filtering for the toy problem can be seen in \figurename~\ref{fig:toy_newPR}. It was computed on a validation set composed of 10 simulated noisy periods. We can see that the post-filtering curve is always above the PR curve of the original model, which means that it performs better. A quantitative comparison for different PR curves and trajectories is proposed in Section~\ref{sec:expe}.

\section{Experimental results}
\label{sec:expe}

\subsection{Experimental setup}

\begin{figure}[!ht]
    \centering
    \includegraphics[width=0.45\textwidth]{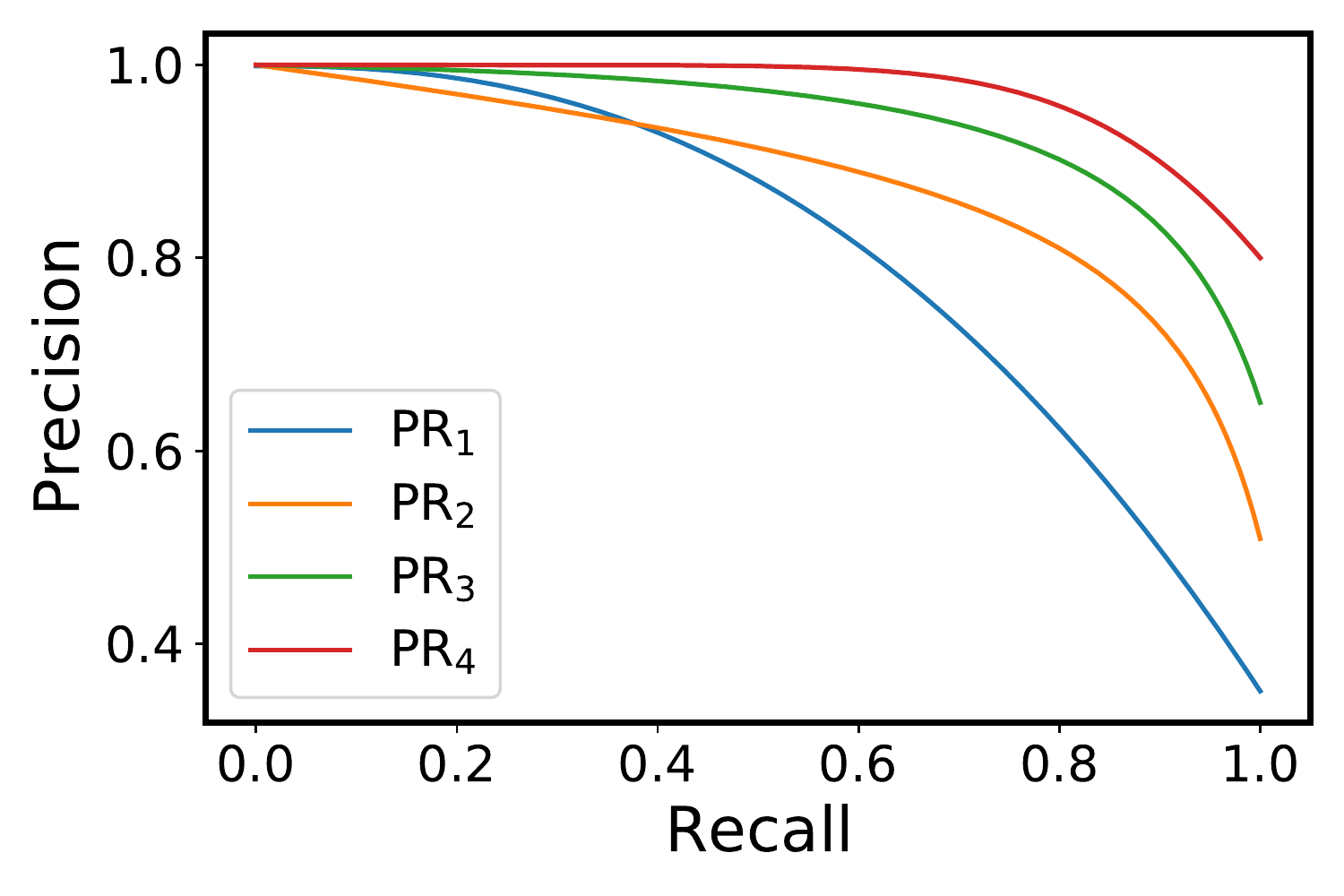}
    \caption{Precision-Recall curves for the 4 simulated OD models used to evaluate our proposed GP filtering approach.}
    \label{fig:eval_PRcurves}
\end{figure}

\begin{figure}
    \centering
    \includegraphics[width=0.45\textwidth]{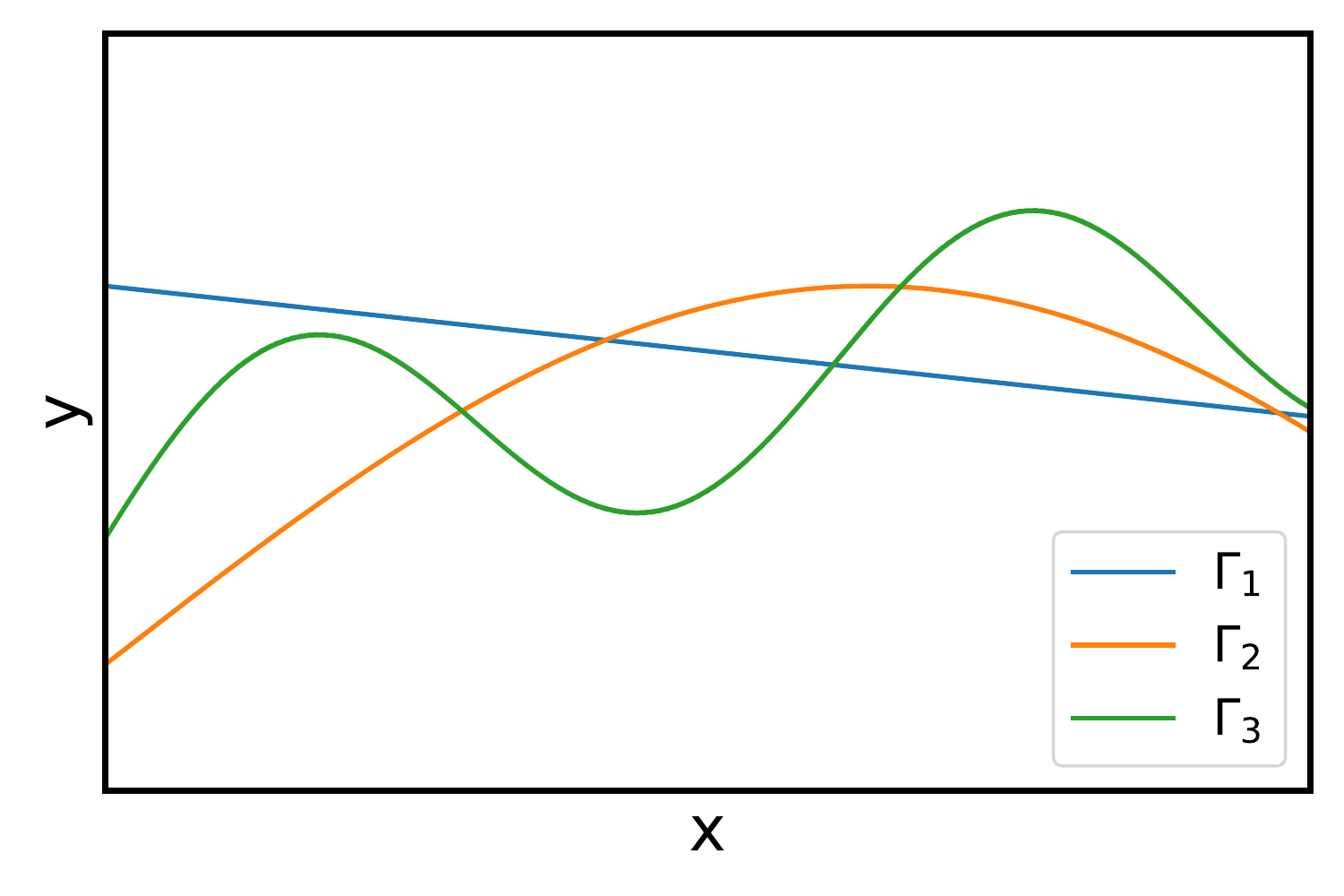}
    \caption{Nominal support lines for the 3 simulated periodic OD trajectories used to evaluate our proposed GP filtering approach.}
    \label{fig:eval_traj}
\end{figure}

In order to test the GP filtering approach presented in Section~\ref{sec:method}, we propose to study the behavior for three different nominal trajectories and four different OD models, characterized by their PR curves. The PR curves of the 4 simulated OD models are shown in \figurename~\ref{fig:eval_PRcurves}. As we can see, they are indexed such that if $i>j$, PR$_i$ is better than PR$_j$. Likewise, \figurename~\ref{fig:eval_traj} represents the nominal trajectory lines in the image plane for all three simulated trajectories used for evaluation. In order to assess the ability of the GP to model different kind of trajectories, we selected one straight line, one convex curve and an oscillating function. For all trajectories, the period $T$ is set to 8 seconds and the time for an object to traverse an image to 5.2 seconds. The evaluation of the proposed method consists in comparing the OD performance before and after filtering, for all 12 possible combinations of trajectories and PR curves. Finally, we note that the toy problem studied all along this paper corresponds to PR$_3$ and $\Gamma_3$.

\subsection{Quantitative Analysis}

As already mentioned, OD models performance can be compared qualitatively by visually analyzing their PR curves. However, in order to conduct a formal objective evaluation, it is better to compare models quantitatively, using relevant numerical metrics. The most commonly used metric to evaluate an OD model is called mean Average Precision. For single class object detection it is simply Average Precision (AP) and it is the mean precision across all possible values of recall. It is defined as the area under the PR curve:
\begin{equation}
    AP = \int_{0}^{1} \text{precision}(r) dr.
\end{equation}

However, it has recently been argued that AP fails to accurately evaluate models for practical scenarios~\cite{olrp}. Indeed, in practice, a model will be used for a single value of the threshold $\tau_s$ and thus at a single point in the PR curve. For this reason, it might be better to evaluate a model by its performance at the optimal PR configuration. Hence, we also introduce the optimal F1-score metric (oF1), which is a variant of oLRP~\cite{olrp}. This new metric is introduced because we do not consider the width and height of the bounding boxes but only their centers in the simulated scenarios studied. In this sense, the \textit{localization} component of oLRP cannot be computed. To define oF1, we first need to introduce the F1-score, which is the harmonic mean of the precision and recall:
\begin{equation}
F1\left(\text{Precision},\text{Recall}\right) = 2 \cdot \frac{\text{Precision} \cdot \text{Recall} }{\text{Precision} + \text{Recall}}.
\end{equation}
From this definition, we can evaluate F1 for all points along the PR curve of a model and choose the precision recall pair with highest F1-score as the operating point for the OD model. This way, the oF1 score of a given model is the value of the F1-score at this operating point. It represents the precision-recall balance of a model in its best configuration and has values in $[0, 1]$. An oF1 score of 1 means that there exist a confidence threshold $\tau_s$ such that the model finds all ground truth BB and does not produce any FP.


The results obtained for the 12 simulated configurations presented above are shown in Table~\ref{tab:results}. For a given metric, the gain in performance by applying the proposed GP filtering approach can be measured by comparing the post-filtering results (Auto cleaning or Manual cleaning) with the reference value of the original model. The bold numbers represent the cases in which there is a gain in performance.

\begin{table}[!ht]
\centering
\caption{Results of the simulated experimental evaluation of our GP filtering approach for periodic OD.}

\begin{subtable}{0.5\textwidth}
\caption*{Average Precision (AP)}
\centering
\begin{tabular}{cc|cccc}
\hline
\multicolumn{2}{c|}{OD model} & PR$_1$ & PR$_2$ & PR$_3$ & PR$_4$ \\ \hline
\multicolumn{2}{c|}{Reference} & 0.811 &	0.883 & 0.942 & 0.975 \\ \hline
\multirow{3}{*}{\parbox{1cm}{\centering Auto cleaning}} & $\Gamma_1$ & 0.483 & 0.606 & \textbf{0.959} & \textbf{0.985} \\
& $\Gamma_2$ & 0.403 & 0.523 & \textbf{0.960} & 0.974 \\
& $\Gamma_3$ & 0.402 & 0.523 & \textbf{0.981} & \textbf{0.986} \\ \hline
\multirow{3}{*}{\parbox{1cm}{\centering Manual cleaning}} & $\Gamma_1$ & \textbf{0.940} & \textbf{0.966} & \textbf{0.959} & \textbf{0.986} \\
& $\Gamma_2$ & \textbf{0.979} & \textbf{0.978} & \textbf{0.988} & \textbf{0.980} \\
& $\Gamma_3$ & \textbf{0.977} & \textbf{0.966} & \textbf{0.974} & \textbf{0.978}\\
\hline
\end{tabular}
\end{subtable}

\vspace{5pt}

\begin{subtable}{0.5\textwidth}
\caption*{Optimal F1-score (oF1)}
\centering
\begin{tabular}{cc|cccc}
\hline
\multicolumn{2}{c|}{OD model} & PR$_1$ & PR$_2$ & PR$_3$ & PR$_4$ \\ \hline
\multicolumn{2}{c|}{Reference} & 0.714 & 0.811 & 0.865 & 0.902
 \\ \hline
\multirow{3}{*}{\parbox{1cm}{\centering Auto cleaning}} & $\Gamma_1$ & 0.542 & 0.749 & \textbf{0.964} & \textbf{0.994} \\
& $\Gamma_2$ & 0.521 & 0.624 & \textbf{0.971} & \textbf{0.992} \\
& $\Gamma_3$ & 0.496 & 0.697 & \textbf{0.987} & \textbf{0.998} \\ \hline
\multirow{3}{*}{\parbox{1cm}{\centering Manual cleaning}} & $\Gamma_1$ & \textbf{0.943} & \textbf{0.983} & \textbf{0.973} & \textbf{0.998} \\
& $\Gamma_2$ & \textbf{0.983} & \textbf{0.990} & \textbf{0.994} & \textbf{0.983} \\
& $\Gamma_3$ & \textbf{0.977} & \textbf{0.986} & \textbf{0.990} & \textbf{0.994}\\ \hline
\end{tabular}
\end{subtable}

\label{tab:results}
\end{table}

\subsection{Discussion}

Table~\ref{tab:results} shows that for all tested OD models and trajectories, if a manual cleaning of training data is performed, the filtering provides an important performance gain, especially for the worst initial models. For example, the original PR$_1$ model does not have any threshold configuration leading F1-score higher than 0.714, whereas after filtering oF1 is higher than 0.94 for all three trajectories. Manual filtering simply means removing the FPs produced by the model in the few trajectories recorded for training, which is not such a tedious and complicated task regarding the potential gain in performance obtained.

Furthermore, we can also see that when the initial model gets better (PR$_3$ and PR$_4$), the cleaning method presented in Section~\ref{sec:method} can automatically handle the training data pre-processing. This means that if we can train a good OD model for a task where the object has a periodic motion, substantial improvement can be obtain with very little human effort, simply by recording a few periods of time and running our automatic method.

\section{Conclusion}
\label{sec:conclusion}

In this paper, we proposed a method to improve robustness of OD models for cases where the object of interest has a periodic motion with respect to the video frames. Our approach consists in gathering a training dataset by running the real system for some periods of time and fitting a Gaussian Process regression model to the observed (x,y,t) positions. This GP is then used at inference time to identify the FPs within the BBs predicted by the OD model. Our approach was tested in simulations for different OD models and trajectories. The obtained results show that this filtering can significantly increase the OD performance for the periodic system.

Several potential directions for future work have been identified. First, the method should be validated on real systems, such as the ones presented in \figurename~\ref{fig:useCase}. Data for this evaluation are currently being gathered by our team to enrich the experimental validation of our approach. Second, more characteristics of the BBs (height, width) could be used as input for the GP filter in order to have more accurate time prediction and by extension more accurate FPs removal. Finally, the approach proposed in this paper for period identification fails for cases where the object never leaves the video frame. Small adaptation can be made in order to succeed in such cases, e.g. running clustering directly in the xy-plane. However, this should be better formalized and tested in the future.


\end{document}